\documentclass[runningheads,a4paper]{llncs}

\usepackage{amssymb}
\usepackage{amsmath}
\usepackage{graphicx}

\usepackage{url}
\urldef{\mailsa}\path|{nccodell, cclin, hindm, rferis, jsmith
}@us.ibm.com|    
\urldef{\mailsb}\path|{halperna
}@mskcc.org|    
\newcommand{\keywords}[1]{\par\addvspace\baselineskip
\noindent\keywordname\enspace\ignorespaces#1}

\newcommand\blfootnote[1]{%
  \begingroup
  \renewcommand\thefootnote{}\footnote{#1}%
  \addtocounter{footnote}{-1}%
  \endgroup
}

\begin{document}

\mainmatter  

\title{Collaborative Human-AI (CHAI): Evidence-Based Interpretable Melanoma Classification in Dermoscopic Images }

\titlerunning{Interpretable Melanoma Classification in Dermoscopic Images }

%
%


\author{Noel C. F. Codella$^1$\blfootnote{Presented at MICCAI 2018, IMIMIC Workshop: https://imimic.bitbucket.io}  \and Chung-Ching Lin$^1$ \and Allan Halpern$^2$ \and Michael Hind$^1$ \and Rogerio Feris$^1$ \and John R. Smith$^1$}
\authorrunning{Codella et al.  }


\institute{ $^1$IBM T.J. Watson Research Center, Yorktown Heights, NY, 10598, USA \\
$^2$Memorial Sloan-Kettering Cancer Center, New York, NY, 10065 \\
$^1$\mailsa\\  
$^2$\mailsb\\
}


%
%

\toctitle{Evidence-Based Interpretable Melanoma Classification in Dermoscopic Images}
\tocauthor{Codella et al.}
\maketitle

\begin{abstract}
Automated dermoscopic image analysis has witnessed rapid growth in diagnostic performance. Yet adoption faces resistance, in part, because no evidence is provided to support decisions. In this work, an approach for evidence-based classification is presented. A feature embedding is learned with CNNs, triplet-loss, and global average pooling, and used to classify via kNN search. Evidence is provided as both the discovered neighbors, as well as localized image regions most relevant to measuring distance between query and neighbors. To ensure that results are relevant in terms of both label accuracy and human visual similarity for any skill level, a novel hierarchical triplet logic is implemented to jointly learn an embedding according to disease labels and non-expert similarity. Results are improved over baselines trained on disease labels alone, as well as standard multiclass loss. Quantitative relevance of results, according to non-expert similarity, as well as localized image regions, are also significantly improved.

\keywords{deep learning, evidence, explainable, interpretable, triplet-loss, global average pooling, weighted activation maps, dermoscopy, melanoma}
\end{abstract}

\section{Introduction}

In the past decade, advancement in computer vision techniques has been facilitated by both large-scale datasets and deep learning approaches. Now this trend is influencing dermoscopic image analysis, where the International Skin Imaging Collaboration (ISIC) has organized a large public repository of high quality annotated images, referred to as the ISIC Archive (http://isic-archive.com). From this repository, snapshots of the dataset have been used to host two consecutive years of benchmark challenges \cite{isbi2017,jaadarticle}, which have increased interest in the computer vision community ~\cite{jaadarticle,codellajrd,nature,recod,montypython}, and supported the development of methods that surpassed the diagnostic performance of expert clinicians \cite{jaadarticle,codellajrd,nature}. However, despite these advancements, deployment to clinical practice remains problematic, in part, because most systems lack evidence for predictions that can be interpreted by users of varying skill. 

Recent works have attempted to provide various forms of evidence for decisions. Methods to visualize feature maps in neural networks were introduced in 2015 ~\cite{visualization}, facilitating better understanding of the behavior of networks, but not justifying predictions made on specific image inputs. Global average pooling approaches have been proposed ~\cite{gap2016}, which get closer to justifying decisions on specific image inputs by indicating importance of image regions to those decisions, but fail to provide specific evidence behind the classifications.

An extensive body of prior work around content-based image retrieval (CBIR) is perhaps the most relevant toward providing classification decisions with evidence ~\cite{cbir1,cbir2,skincbir,cbir3,nipscbir}. Early approaches relied on low-level features and bag-of-visual words,  ~\cite{cbir1,cbir2,skincbir}, but suffered from the ``semantic gap'': feature similarity did not necessarily correlate to label similarity.  Later approaches have used deep neural networks to learn an embedding for search, reducing semanic gap issues  ~\cite{nipscbir}. However, such methods have still suffered from a ``user-gap'': what an embedding learns to consider as similar from disease point-of-view does not necessarily correlate with human measures of similarity. In addition, users cannot determine what spatial regions of images contributed most to distance measures. 

Specific to the domain of dermoscopic image analysis, one work proposed to learn and localize clinically discriminative patterns in images ~\cite{recod}; however, this output can only be verified by experts who know how to identify the patterns.  In addition, classifier decision localization has been proposed for multimodal systems ~\cite{zongyuan}; however, localization information alone isn't sufficient as evidence for classification decisions.

In this work, a solution for a Collaborative Human-AI (CHAI) dermoscopic image analysis system is presented. In order to facilitate interpretability of evidence by clinical staff of any skill level, this approach 1) introduces a novel hierarchical triplet loss to learn an embedding for k-nearest neighbor search, optimized jointly from disease labels as well as non-expert human similarity, and 2) provides localization information in the form of {\em query-result activation map pairs}, which designate regions in query and result images used to measure distance between the two. Experiments demonstrate that the proposed approach improves classification performance in comparison to models trained on disease labels alone, as well as models trained with classification loss. The relevancy of results, according to non-expert similarity, are also significantly improved.

\section{Methods}
\label{methods}

\subsection{Triplet-Loss with Global Average Pooling}

\begin{figure}[t]
  \centerline{\includegraphics[width=12cm]{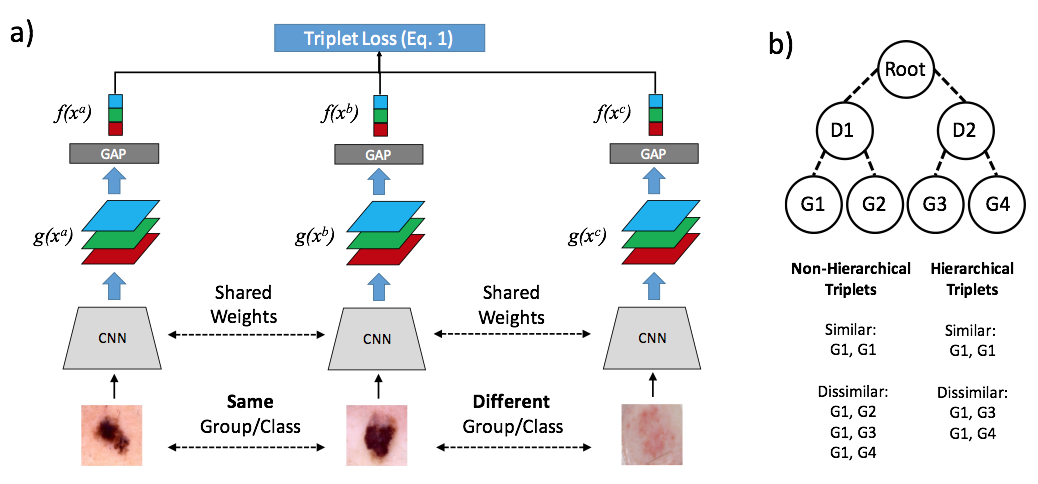}}
  \caption{ a) Proposed triplet loss framework with global average pooling (GAP) architecture. b) Top: Visual example of proposed hierarchical annotation groups. The first level grouping is by disease label (D1-2), and the second level by human visual similarity (G1-4). Bottom: Example triplet logic is shown as pairing between groups. }
\label{fig:arch}
\end{figure}

The proposed embedding framework is displayed in Fig. ~\ref{fig:arch}a. A triplet loss structure  ~\cite{tripletloss} is combined with penultimate global average pooling layers ~\cite{gap2016}  to learn a discriminative feature embedding that supports activation localization. AlexNet, including up to the ``conv5'' layer, is used as the CNN. 

In order to train, 3 deep neural networks with shared weights across 3 input images $(x^{a},x^{b},x^{c})$ produce feature embeddings $(f(x^{a}),f(x^{b}),f(x^{c}))$. The following objective function over those embeddings provides the gradient for backpropagation:

\begin{equation}
L=max\left [ 0,l+D(f(x^{a}),f(x^{b})) - \frac{1}{2}(D(f(x^{a}),f(x^{c})) + D(f(x^{b}),f(x^{c})))  \right ]
\end{equation}

where $D()$ is a distance metric (squared Euclidean distance), $l$ is a constant representing the margin (set to 1), $x^a$ and $x^b$ are considered similar inputs, and $x^c$ is a dissimilar input. 

The feature embedding is comprised of a global average pooling (GAP) layer to support generation of a {\em query-result activation map pair}, which highlights regions of pairs of images that contributed most toward the distance measure between them. This is done by combining the feature layer activation maps prior to global average pooling into a single grayscale image, weighted by the squared differences between two image feature embeddings: 

\begin{equation}
A^{q}(i,j)=\sum_{z=0}^{d}g_{z}(x^{q},i,j)) \cdot (f_{z}(x^{q})-f_{z}(x^{r}))^{2}
\end{equation}

where $A^{q}(i,j)$ is the query activation map (QAM), $g_{z}(x,i,j)$ is the $z^{th}$ filter bank before global average pooling, $d$ is the dimensionality of the filter bank, $x^{q}$ is the query image, $x^{r}$ is a search result image, and:

\begin{equation}
f_{z}(x) = \frac{1}{n^{2}} \sum_{i=0}^{n}\sum_{j=0}^{n}g_{z}(x, i, j) 
\end{equation}

is the $z^{th}$ feature embedding element. The result activation map (RAM) $A^{r}$ in the query-result pair is likewise computed as in Eq. 2, where $g_{z}(x^{q},i,j)$ is replaced with $g_{z}(x^{r},i,j)$. 


\subsection{Hierarchical Triplet Selection Logic}

An example of the hierarchical triplet selection logic is shown in Fig. ~\ref{fig:arch}b. Given visually similar groups annotated under disease labels, a hierarchical selection process pairs images as similar if they are siblings within the same group under a disease parent. Dissimilar images include images from other disease states, but exclude cousin images (images within the same disease, but different similarity group). A non-hierarchical selection process takes dissimilar images from any other group, including cousins. 

\begin{figure}[t]
  \centerline{\includegraphics[width=12cm]{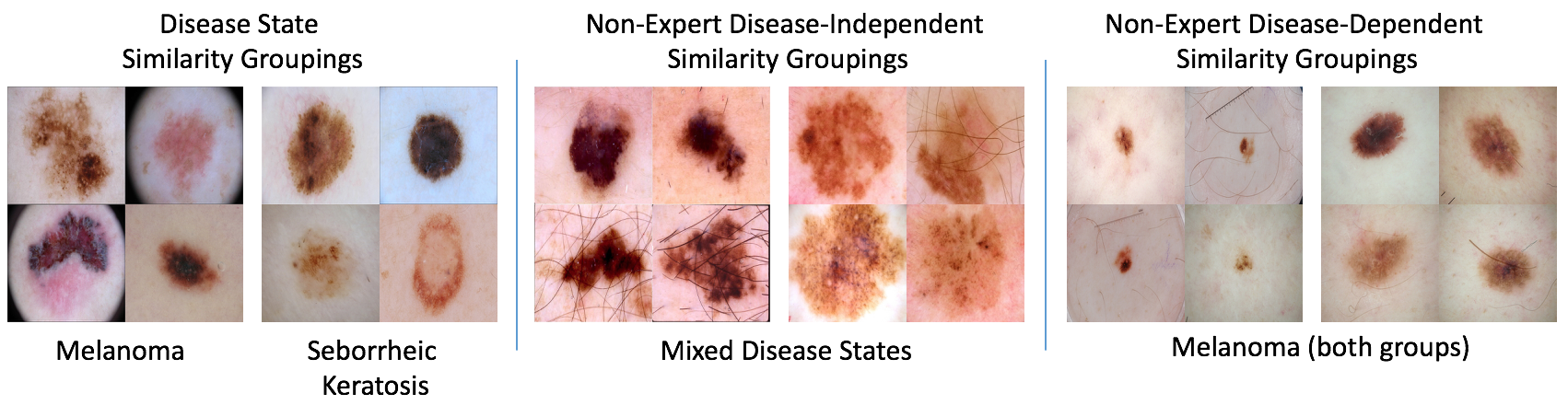}}
  \caption{ Example groups by disease category from the ISIC database (left), by non-expert similarity disregarding disease diagnosis (center), and by non-expert similarity constrained within disease groups (right).  }
\label{fig:groups}
\end{figure}

\subsection{Experimental Design}

The 2017 International Skin Imaging Collaboration (ISIC) challenge on Skin Lesion Analysis Toward Melanoma Detection ~\cite{isbi2017} dataset is used for experimentation. This is a public dataset consisting of 2000 training dermoscopic images and 600 test images. Experiments on this data compare between the following 6 feature embeddings for kNN classification: 

{\bf Baseline: } The first is the 4096 dimensional fc6 feature embedding layer of the AlexNet architecture trained on the CASIA-WebFace dataset, described in prior work~\cite{tripletloss}. This is used as the baseline as it is one of the only human-skin focused pre-trained networks currently available. 

{\bf BaselineFT: } Baseline 4096 is fine-tuned for disease labels using standard multiclass accuracy loss. This method represents one of the most common approaches for generating embeddings for KNN classification in practice. 

{\bf Disease: } This is a 1024 dimensional CHAI feature embedding, learned from disease labels on the training data partition of the ISIC dataset, fine-tuned from the baseline. 

{\bf Joint: } This is a CHAI feature embedding jointly fine-tuned from baseline using disease labels, as well as non-expert human similarity groupings, consisting of 1700 images pulled from the ISIC Archive (excluding test images), annotated into 37 distinct groups. The annotator was not given disease labels, and thus may mix diseases within groups.  Example groups are shown in Fig. ~\ref{fig:groups}.

{\bf Hierarchical: } This is a CHAI feature embedding fine-tuned from the disease model using human similarity groups that are dependent on disease labels. All 2000 images and 600 test images were annotated from the 2017 ISIC challenge dataset, partitioned into 20 groups of similar images under melanoma, 12 groups under seborrheic keratosis, and 15 groups under benign nevus, according to a non-expert human user.  Because this type of data is difficult to annotate, only 1000 training images were used for fine-tuning. The remainder of the data was used for evaluation. Examples of these groups are shown in Fig. ~\ref{fig:groups}. Triplets were selected based on hierarchical logic.

{\bf Non-Hierarchical: } To isolate the effects of hierarchical logic, and disease labels being provided to the annotator, the hierarchical groups are used to create triplets using non-hierarchical logic: dissimilar images are selected from any other group, including cousins.  

\begin{figure}[t!]
  \centerline{\includegraphics[width=12cm]{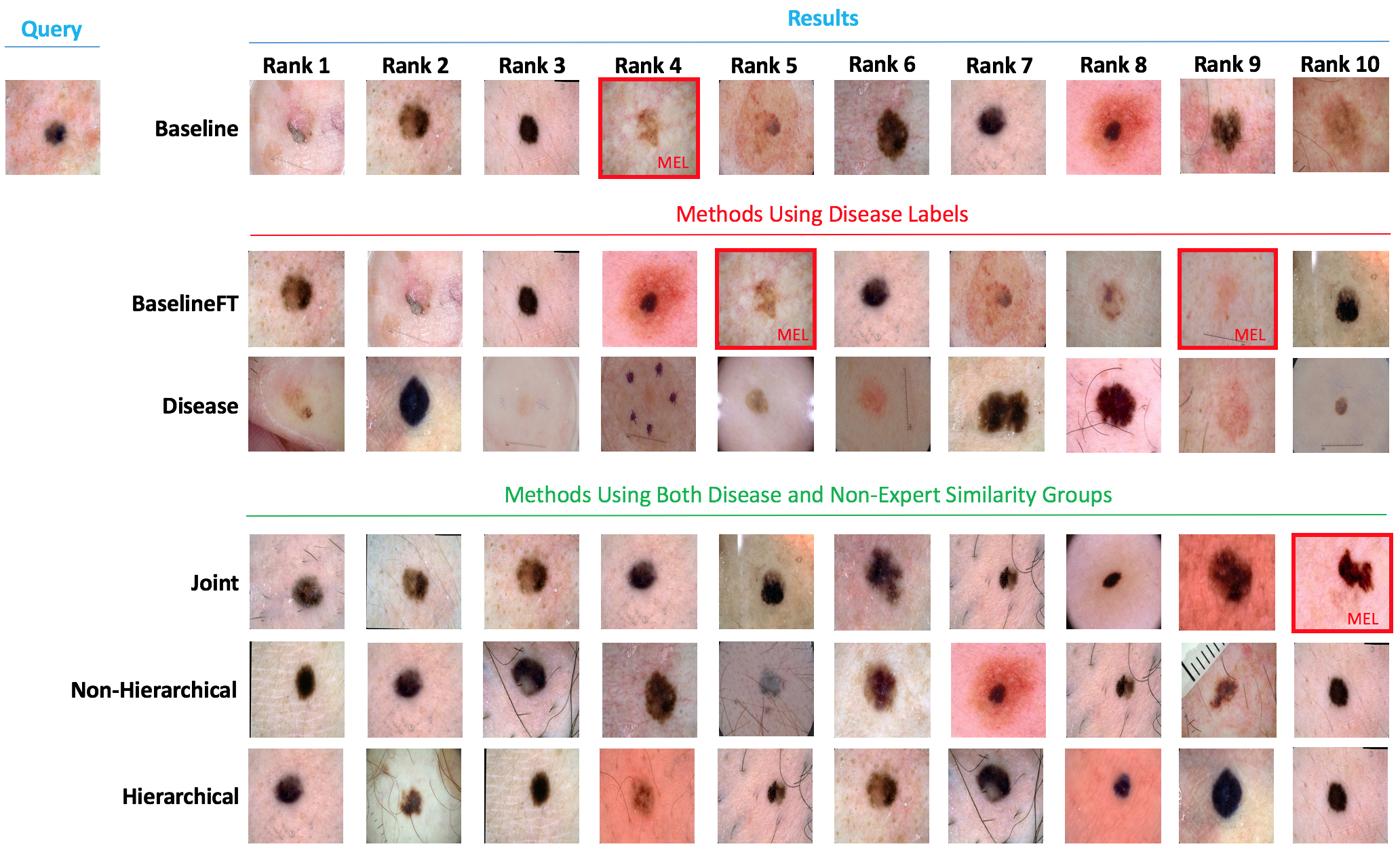}}
  \caption{ Example search results across systems, displayed according to similarity rank, with rank 1 being the most similar image in the training dataset. Red borders signify instances of melanoma. }
\label{fig:search}
\end{figure}

Most learning parameters are kept consistent with prior art ~\cite{tripletloss}, including the activation map feature dimensionality of 1024 ~\cite{gap2016}, batch size 128, momentum of 0.9, ``step'' learning rate policy, learning rate for transferred weights (0.00001), and learning rate for randomly initialized GAP layer (0.01). For BaselineFT, a learning rate of 0.01 was used for fc8, 0.001 for fc7 and fc6, and 0.00001 for earlier layers. For all triplet experiments, 150,000 triplets were randomly generated for training, and 50,000 triplets for validation.

The area under receiver operating characteristic (ROC) curve (AUC) is used to measure melanoma classification performance on the dataset, according to average vote among returned nearest neighbors. The hierarchical similarity annotations were used to measure the average number of results matching non-expert human relevancy (REL) across all experiments. Finally, the quality of query activation maps are quantitatively measured by comparing the maps against ground truth segmentation according to Jaccard (JA).

\section{Results}
\label{results}

\begin{table}[t!]
\centering
\begin{center}
\begin{tabular}{ |c|c|c|c|c|c|c| } 
 \hline
 {\bf } & {\bf Baseline} & {\bf BaselineFT} & {\bf Disease} & {\bf Joint}  & {\bf Non-Hierarchical} & {\bf Hierarchical} \\ 
 \hline
 {\bf AUC k3} & 0.663 & 0.700 & {\bf 0.734} & 0.704  & 0.713 & 0.729 \\ 
 {\bf AUC k5} & 0.675 & 0.714 & 0.744 & 0.738  & 0.743 & {\bf 0.756} \\ 
 {\bf AUC k10} & 0.681 & 0.709 & 0.757 & 0.754  & 0.749 & {\bf 0.774}\\ 
 {\bf AUC k20} & 0.712 & 0.745 & 0.775 & 0.752  & 0.769 & {\bf 0.783} \\ 
 {\bf AUC k40} & 0.691 & 0.742 & 0.776 & 0.760  & 0.776 & {\bf 0.786} \\ 
  \hline
 {\bf REL k3} & 0.942 & 1.005 & 0.865 & 1.048  & {\bf 1.212} & 1.125 \\
{\bf REL k5} & 1.505 & 1.608 & 1.412 & 1.678  & {\bf 1.958} & 1.872 \\
{\bf REL k10} & 2.875 & 3.027 & 2.632 & 3.147  & {\bf 3.793} & 3.658 \\ 
{\bf REL k20} & 5.470 & 5.772 & 4.903 & 6.067  & {\bf 7.300} & 6.968 \\
{\bf REL k40} & 10.283 & 10.703 & 9.125 & 11.507 & {\bf 13.958}  & 13.333 \\ 
  \hline
 {\bf JA} & NA & NA & 0.176 & 0.201  & 0.193 & {\bf 0.208} \\
 \hline
\end{tabular}
 \caption{Melanoma Classification AUC for each method and number of neighbors (k), followed by number of results matching human similarity relevancy (REL), and Jaccard (JA) of QAM against segmentation ground truth. }
\label{table:auc}
\end{center}
\end{table}

Table ~\ref{table:auc} shows the measured AUC for each model type and variable number of neighbors (k), the number of results matching non-expert human similarity relevance (REL), and the Jaccard of the query activation maps as judged against ground truth segmentations. For comparison, standard classification output from multi-class loss used to train {\em BaselineFT} produces an AUC of 0.772. The top AUC measured for the challenge was 0.874 ~\cite{recod}.

For $k=3$, {\em Disease} achieved the highest AUC. Surprisingly, at $k=20,40$, {\em Disease} outperforms the classification output of {\em BaselineFT} (0.772 AUC). For all other values of $k$, the {\em Hierarchical} triplet loss embedding achieved the highest performance. At $k=40$, these performance numbers were comparable with predictive systems submitted to the challenge (rank 11 out of 23 submissions). The {\em Hierarchical} triplet loss also achieved the second highest number of human similarity relevant results. While the {\em Non-Hierarchical} method achieved the highest degree of human similarity relevant results, this came at the marginal cost of some classification performance in comparison to {\em Hierarchical} triplets. However,  {\em Non-Hierarchical} has still matched the classification performance of {\em Disease}, and outperformed the standard multiclass loss of {\em BaselineFT}. {\em Joint} also showed improvements to relevance of human similarity in comparison to {\em Disease}, but suffered a more harsh penalty to classification performance in comparison to {\em Hierarchical} and {\em Non-Hierarchical}. 

Representative search results can be inspected in Fig. ~\ref{fig:search}. One can observe here how {\em Disease}, trained directly on triplets from disease labels,  does not translate into the most ``relevant'' results by human measure: clearly, rank 3 has returned a hypo-pigmented lesion for a pigmented lesion query. In contrast, {\em Joint}, while maintaining a robust improvement in AUC measures over {\em Baseline} and {\em BaselineFT}, has additionally learned to balance disease similarity with a more human measure of similarity. {\em Hierarchical} has both managed to improve classification performance and human similarity.

\begin{figure}[t]
  \centerline{\includegraphics[width=11cm]{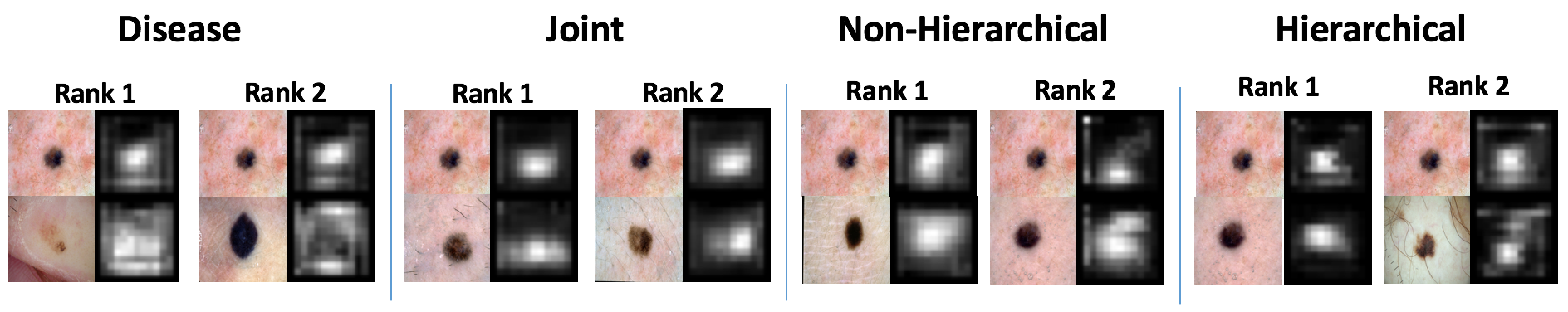}}
  \caption{ Example query-result activation pairs for search results. In each group of 4 images: {\em Top-Left:} query image. {\em Top-Right:} query activation map. {\em Bottom-Left:} search result.  {\em Bottom-Right:} search result activation map. }
\label{fig:localization}
\end{figure}

Finally, example query-result activation map pairs are shown in Fig. ~\ref{fig:localization}. Interestingly, {\em Disease}  learned to examine a broad image extent during comparisons (even potentially irrelevant areas of images), whereas for the models trained with human measures of similarity, the systems have learned to focus more to the localized lesion area. This is confirmed in the over 10\% quantitative improvement in Jaccard index comparing to ground truth lesion segmentations, as shown in Table ~\ref{table:auc}. 

\section{Conclusion}
\label{conclusion}

In conclusion, ``CHAI'', a Collaborative Human-AI system to perform comprehensive evidence-based melanoma classification in dermoscopic images has been presented. Evidence is provided as both the nearest neighbors used for classification, as well as query-result activation map pairs that visualize regions of the images contributing most toward a distance computation. Using a novel hierarchical triplet loss, non-expert human similarity is used to tailor the feature embedding to more closely approximate human judgments of relevance, while simultaneously improving classification performance and the quality of the activation maps. Future work must be carried-out to determine 1) whether the method has the potential to improve adoption, 2) how to improve classification performance to better compete with other black-box systems, and 3) whether passive user interaction with a deployed system can be used for training (for example, from a user clicking on a specific evidence result) to improve classification performance and relevance over time with continued use.

\pagebreak

\end{document}